\documentclass[runningheads]{llncs}
\usepackage{graphicx}
\usepackage{amsfonts}
\usepackage{hyperref}
\hypersetup{colorlinks=true}
\begin{document}
\title{SERCNN: Stacked Embedding Recurrent Convolutional Neural Network in Detecting Depression on Twitter}
\titlerunning{SERCNN in Detecting Depression on Twitter}
%
\author{Heng Ee Tay\inst{1}\orcidID{0000-0002-8182-3775} \and
Mei Kuan Lim\inst{1}\orcidID{0000-0001-8834-9933} \and
Chun Yong Chong\inst{1}\orcidID{0000-0003-1164-0049}}
\authorrunning{H.E. Tay et al.}
%
\institute{Monash University Malaysia, 47500 Selangor, Malaysia 
\email{\{Heng.Tay,Lim.MeiKuan,Chong.ChunYong\}@monash.edu}}

\maketitle              
\begin{abstract}
Conventional approaches to identify depression are not scalable and the public has limited awareness of mental health, especially in developing countries. As evident by recent studies, social media has the potential to complement mental health screening on a greater scale. The vast amount of first-person narrative posts in chronological order can provide insights into one's thoughts, feelings, behavior, or mood for some time, enabling a better understanding of depression symptoms reflected in the online space. In this paper, we propose SERCNN, which improves the user representation by (1) stacking two pretrained embeddings from different domains and (2) reintroducing the embedding context to the MLP classifier. Our SERCNN shows great performance over state-of-the-art and other baselines, achieving 93.7\% accuracy in a 5-fold cross-validation setting. 
Since not all users share the same level of online activity, we introduced the concept of a fixed observation window that quantifies the observation period in a predefined number of posts. With as minimal as 10 posts per user, SERCNN performed exceptionally well with an 87\% accuracy, which is on par with the BERT model, while having 98\% less in the number of parameters. Our findings open up a promising direction for detecting depression on social media with a smaller number of posts for inference, toward creating solutions for a cost-effective and timely intervention. We hope that our work can bring this research area closer to real-world adoption in existing clinical practice.

\end{abstract}
\section{Introduction}
Depression is a serious, yet common mental disorder that affects more than 264 million people worldwide. The number is projected to grow amid the war against the pandemic of COVID-19. Unlike the usual mood fluctuations and emotional responses, long-lasting sadness, emptiness, or irritation in one's day-to-day life is usually accompanied by somatic and cognitive changes that heavily disrupt an individual's functioning capacity. Hence, depression is often associated with suicide at its worst. In the latest Word Health Statistics 2021 \cite{world2021world}, there is a 28\% increase in the suicide rate in the United States in this period.

Often, people who suffer from depression tend to be unaware of their mental condition, as there is no specific feedback from the body, like common physical injuries do. Furthermore, existing depression screening approaches through self-reporting questionnaires and clinical interviews are expensive and not scalable. Currently, many countries, especially developing countries, are suffering from a shortage of provisions and services to identify, support, and treat mental health issues. Therefore, there is a dire need to promote mental health awareness and create technologies to help complement existing screening or diagnosis approaches to help reach a wider community. Patients are required to disclose truthful information during clinical interviews \cite{dsm5}. Until now, global provision and services to identify, support, and treat mental health problems are insufficient despite the disruption of essential health services, community mistrust, and fears of COVID-19 infections.

Meanwhile, social media has become part and parcel of everyday life. The availability of a large volume of data has motivated research to utilize social media data for depression detection. Social media data can serve as a life log that records user activities in text, image, audio, or video. In addition, the enormous amount of first-person narrative posts can provide insight into one's thoughts, behavior, or mood in the temporal dimension. 

Previous studies 
have demonstrated the great potential of social media posts in detecting mental disorders, such as depression \cite{coppersmith2015clpsych,holleran2010early,losada2017erisk}, anxiety \cite{GRUDA2019245,guntuku2019twitter}, and eating disorders \cite{yan2019automatic}. In existing depression diagnostics using DSM-5 \cite{dsm5}, a 14-day observation period is required. However, using machine learning-based models, Hu et al.\cite{7396831} and Tsugawa et al. \cite{10.1145/2702123.2702280} have suggested that two months of posting data are used to detect depression. Lately, Shen et al. \cite{shen2017depression} have constructed a depression dataset from users' one-month posting history. Shen et al. \cite{shen2017depression} and Gui et al. \cite{gui2019depression,gui2019cooperative} have demonstrated that a single month of posting history is sufficient to classify depression online. However, there is still a lack of evidence on the number of posts required for efficient and cost-effective detection in the online space.

In this paper, we propose SERCNN - Stacked Embedding Recurrent Convolutional Neural Network, to perform early depression detection based on their one-month social media posts. Our SERCNN improves the overall user representations in two ways: (1) stacking representation vectors from pretrained embeddings from both the general and social media domains and (2) allowing the MLP classifier to exploit the context by co-learning the CNN context with the recurrence of the embedding context. 
We also trained SERCNN on different lengths of posting history measured in the number of posts (fixed observation windows) to offer a unique perspective on optimizing the amount of data required for early depression detection. We further assessed the robustness of the observation window by comparing the performance differences trained on the subsets extracted from the earliest (head) and latest (tail) of a user-posting history. Despite having a fraction of the BERT \cite{devlin-etal-2019-bert} classifier's size (in terms of the number of parameters), SERCNN is on par, if not better than our BERT baseline, confirming that depression can be identified even if we have only 10 posts for a given user. This discovery sheds light on the adoption of social media analytics to complement existing clinical screening practices without a substantial investment in computational resources.

The contributions of this paper can be summarized below:
\begin{enumerate}

    \item We show that it is possible to improve the feature representation of social media text by stacking features from two low-dimensional pretrained embeddings. As a result, the number of feature dimensions is also reduced.
    
    
    \item We propose the concept of a fixed observation window that quantifies the observation period in a predefined number of posts and provide insights into the potential for early depression detection using the observation window with SERCNN.
    
    \item We highlight the performance and advantage of SERCNN over the fine-tuned BERT classifier in early depression detection for its fraction of BERT's size in terms of number of parameters.

\end{enumerate}

The rest of the paper is organized as follows: Section \ref{sec:related} covers the related work in this domain and Section \ref{sec:motivation} describes the problem statement and motivation of this research. The dataset used in the experiments is elaborated in Section \ref{sec:dataset} and the proposed method is discussed in Section \ref{sec:proposed}. In Section \ref{sec:experiments}, we present the experiment settings and discuss the experimental results in Section \ref{sec:results}. Finally, Section \ref{sec:conclusion} summarizes the main findings of this work and some perspectives for future research.

\section{Related Work} \label{sec:related}

Various natural language processing (NLP) techniques have been applied to extract relevant features from social media data. Modeling feature representations is a crucial task in machine learning; features that are not discriminative will result in poor and faulty model performance. Hence, earlier research works are mainly focused on feature extraction techniques. Choudhury et al. \cite{Choudhury2013PredictingDV} and Tsugawa et al. \cite{10.1145/2702123.2702280} have found that depressed users tend to be emotional. \cite{wang2013depression} have found that using sentiment analysis in depression detection can achieve about 80\% accuracy. Resnik et al. \cite{resnik2015beyond} extracted topics distribution with Latent Dirichlet Allocation (LDA) \cite{10.5555/944919.944937} to differentiate depressed individuals from the healthy controls. Researchers have also extracted features based on the industry standard, the Diagnostic and Statistical Manual of Mental Disorders (DSM), such as the insomnia index derived from the user posting time \cite{shen2017depression}. Linguistic Inquiry and Word Count (LIWC) \cite{LIWC} is a widely used word matching-based feature extraction tool that builds on top of previous findings \cite{gortner2006benefits,SMT3} discovered decades ago. Both Choudhury et al. \cite{Choudhury2013PredictingDV} and \cite{shen2017depression} found that depressed users tend to have a high focus on self-attention, increased medical concerns, and increased expression of religious thoughts. These findings are consistent with previous work \cite{SMT3}, in which depressive indicators are identifiable from human-generated content. Furthermore, Shen et al. \cite{shen2017depression} have also attempted to identify sparse features of their multimodal handcrafted features with dictionary learning. Their studies have discovered that depressed users tend to post tweets at midnight, which could be a result of insomnia. Furthermore, depressed users tend to use emotional words and depression-related terms (antidepressants and symptoms) more often than the controls.

However, the aforementioned feature extraction approaches are often ineffective due to the nature of noisy and unstructured social media text.  An extensive data cleaning and preprocessing are required for the feature extraction to work as intended. Recent advancements in deep learning have shown its robustness without needing manual annotation and label supervision. Word embedding approaches are introduced to learn disentangled representations by learning a mapping function that takes all the embeddings as input and outputs a compact representation of the words. Deep neural networks, such as the recurrent neural network (RNN) and convolutional neural network (CNN) \cite{https://doi.org/10.48550/arxiv.1408.5882}, can further leverage and exploit these disentangled text representations to reach to state-of-the-art performance. The learned representation can be used as a discriminative classifier for downstream tasks. With the policy gradient agent implemented alongside, authors in \cite{gui2019depression} have leveraged word embedding learned from the depression dataset and deep learning models (RNN and CNN) to achieve state-of-the-art depression detection results. These works have shown that deep learning models can outperform shallow models in accuracy and sensitivity with minimal preprocessing effort.

\section{Problem Statement and Motivation} \label{sec:motivation}

Depression in developing countries has always been under-addressed. Many NGOs, local governments, and clinics in these countries do not have ample computing power resources to train and deploy a fully automated screening model that utilizes cutting-edge deep learning, such as BERT. Therefore, there is a dire need for a cost-effective, efficient, and robust early depression detection solution.

\section{Depression Dataset} \label{sec:dataset}

We train and evaluate our proposed method using the Shen Depression Dataset \cite{shen2017depression}, D, which comprises two labeled user groups: (D$_1$) self-declared depressed users and (D$_2$) controls. We formed a balanced dataset by drawing 1,401 samples from D$_1$ and D$_2$, resulting in 2,802 samples and 1.35 million posts made within a month to eliminate bias in our results which may be caused by an imbalanced dataset. On average, a depressed user tends to post lesser at 165 posts as compared to 799 posts for the control user. An overview of the dataset is presented in Table \ref{tab:tab-dataset1}. 

\begin{table}[htb!]
\centering
\caption{Overview of the dataset}
\label{tab:tab-dataset1}
\begin{tabular}{lccc}
\hline\hline
\textbf{Dataset} & {\textbf{Number of users}} & {\textbf{Number of posts}} & {\textbf{Average posts}}\\ \hline\hline
D$_1$ Depressed             & 1,401             & 231,494  & ~165         \\
D$_2$ Controls         & 1,401             & 1,119,080 & ~799      \\ 
\hline
D (D$_1$ + D$_2$) &{2,802 }&{1,350,574} & {~482} \\
\hline\hline

\end{tabular}

\end{table}

We have also performed a statistical analysis on the dataset to better understand how the majority of the users behave online in terms of the posting frequency, which is displayed in Table \ref{tab:tab-dataset-stat}. We can observe that 75\% of the users in the dataset have at least 34 posts (first quartile) and a midspread of more than 400 posts (interquartile range). The dataset also has users with only one post; the longest posting history is 11,127.

\begin{table}[htb!]
\centering
\caption{Statistics of the number of posts per user}
\label{tab:tab-dataset-stat}
\begin{tabular}{lc}
\hline\hline
\textbf{Statistics}       & \textbf{Number of posts}     \\\hline\hline
First quartile     & 49    \\
Median & 154   \\
Third quartile     & 479   \\
Interquartile range    & 430  \\ \hline
Minimum    & 1          \\
Maximum    & 11127  \\ \hline\hline
\end{tabular}%
\end{table}

\section{Proposed Method} \label{sec:proposed}

\begin{figure*}[htb!]
    \centering
    \includegraphics[width=1.03\linewidth]{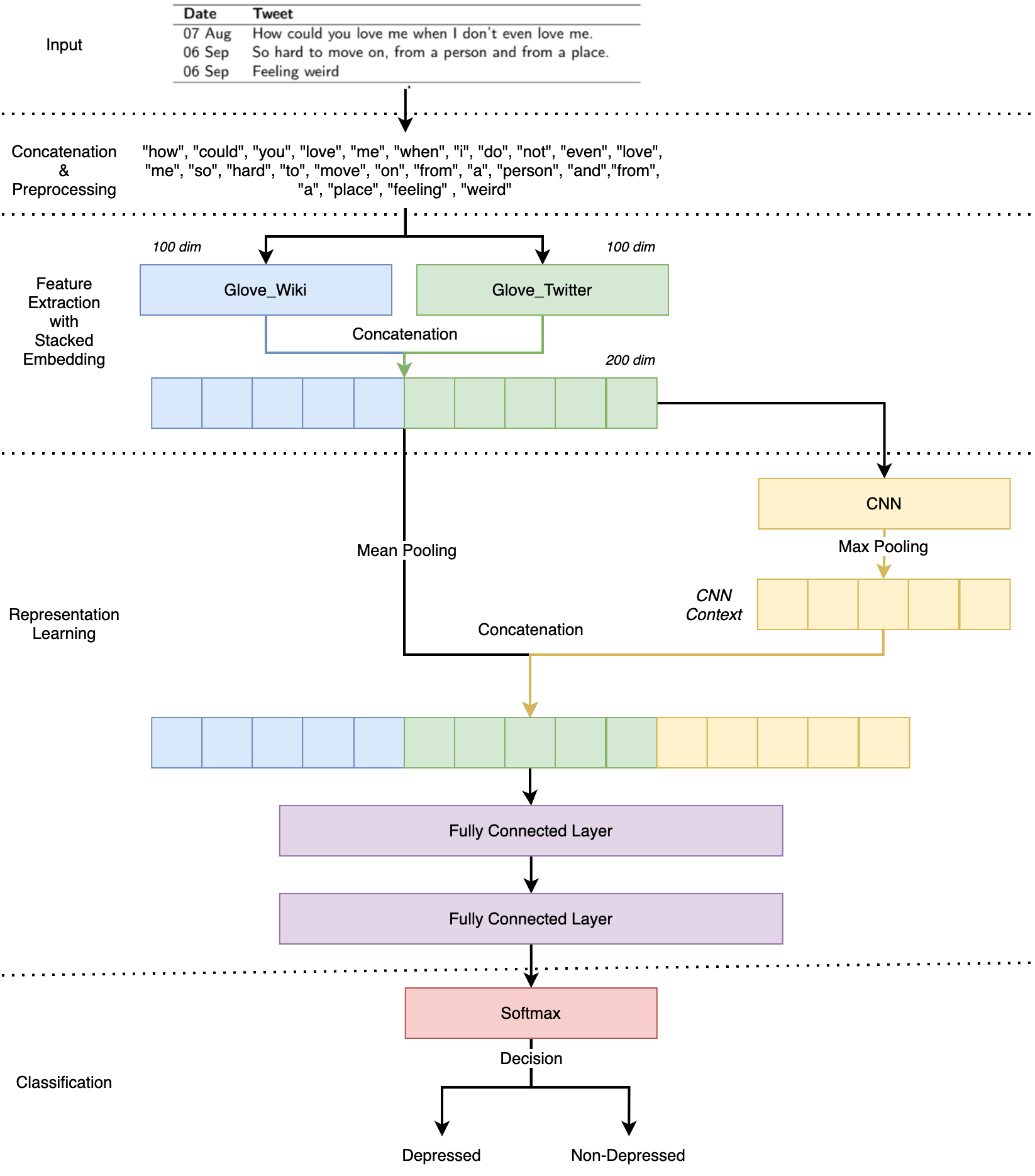}
    \caption{The overall architecture of SERCNN. }
    \label{fig:sercnn}
\end{figure*}

\subsection{Background}

Given that our dataset, $\mathcal{D}$, consists of $\textnormal{N}_1$ number of social media users $\mathbf{u}$, where each user's $\textnormal{N}_2$ number of social media posts, $\mathbf{p}$, within a month were collected, and each post has $\textnormal{N}_3$ words, $w$, we denote the dataset as $\mathcal{D} = \{(u_1, y_1), ..., (u_{\textnormal{N}_1}, y_{\textnormal{N}_1})\}$, where $y \in [0,1]$ is the label in which 0 represents the \textit{control user} and 1 for the \textit{depressed user}. For a random social media user $u_i$, it can be formulated as $u_i = \{p^i_1, ... , p^i_{\textnormal{N}_2}\}$, and a random social media post as ${p_j} = \{w^j_1, ..., w^j_{\textnormal{N}_3}\}$.

\subsection{SERCNN}\label{pmrcnn}

We propose a Stacked Embedding Recurrent Convolutional Neural Network (SERCNN), which is made up of a stacked embedding (SE) and recurrent convolutional neural network (RCNN) \cite{rcnn}. The overall architecture of SERCNN is simple, consisting of a SE layer, TextCNN \cite{https://doi.org/10.48550/arxiv.1408.5882}, a mean-pooling layer, and two fully connected layers, with the latter one being as an output layer with Softmax function, as visualized in Fig. \ref{fig:sercnn}.

For the input of the SERCNN, we model our user representation by first concatenating $N_2$ social media posts in chronological order. For a random user $u_i$, it can be formulated as (padded where necessary):
\begin{equation}
u_i = \{p^i_1 \oplus ... \oplus p^i_{\textnormal{N}_2}\}
\end{equation}
\begin{equation}
u_i = \{w^1_1\oplus ... \oplus w^{1}_{N_3}\oplus ... \oplus w^{N_2}_{N_3}\}
\end{equation}
where $\oplus$ is the concatenation operator. 

The concatenated text can be viewed as a single unique ``monthly diary", journal, or large document that characterizes the user. By concatenating the user-generated text, it allows the deep learning model to take advantage of the context across different time frames (global context) without losing much information throughout the time span.

Then, we extract the distributed text representation of each word with Stacked Embedding (SE); an ensemble embedding technique commonly falls under the category of Meta-embedding. 
The concept of Meta-embedding was first introduced by \cite{yin2016learning} to utilize and learn the meta of existing well-trained pretrained embeddings and extend the vocabulary. Meta-embedding formed from pre-trained embeddings trained on heterogeneous datasets allows for an improved vocabulary coverage and reduces out-of-vocabulary words. As the name suggests, SE is formed by stacking a collection of dense vectors (pretrained weights) $\mathbf{E} = \{\mathbf{E}_1, ... \\, \mathbf{E}_{\textnormal{N}_4}\}$ extracted from $\textnormal{N}_4$ number of pretrained embeddings trained on heterogeneous datasets. SE can be formulated with:
\begin{equation}
    \mathbf{E}_{\textnormal{SE}} = \mathbf{E}_1 \oplus ... \oplus \mathbf{E}_{\textnormal{N}_4}
\end{equation}
where the embedding context, $\mathbf{c}_{\textnormal{SE}}$, of a given user $u$ can be obtained via:
\begin{equation}
     \mathbf{c}_{\textnormal{SE}} = \mathbf{E}_{\textnormal{SE}}(u)
\end{equation}

The vocabulary of the stacked embedding, $\mathbf{V}_{\textnormal{SE}}$, is now considered as the union of the $N_4$ pretrained embeddings' vocabularies, resulting in an extensive vocabulary than a single domain embedding as shown in formula \ref{eq:vocab}.

\begin{equation}
    \mathbf{V}_{\textnormal{SE}}= \bigcup\limits_{n=1}^{\textnormal{N}_4} \mathbf{V}_{n} 
    \label{eq:vocab}
\end{equation}


In this study, the following pre-trained embeddings are used:

\begin{enumerate}

 \item \textbf{{Glove Twitter (100 dimensions)}} is trained using global word co-occurrences information by Pennington et al. \cite{pennington2014glove} under an uncased setting, using 2 billion tweets, with 27 billion tokens. The resulting vector consists of 1.2 million vocabularies learned from the corpus.

 \item \textbf{{Glove Wikipedia 2014 + Gigaword 5 (100 dimensions)}}, similarly, Pennington et al. \cite{pennington2014glove} have pretrained a word embedding from the corpus made up of Wikipedia 2014 and Gigaword 5 datasets using their proposed global word co-occurrences information. The embedding covers approximately 400 thousand words.

\end{enumerate}

The recurrent neural network (RNN) is capable of capturing contextual information over a long sequence. However, the RNN model favors the latter words over words in the earlier sequence. 
Since we are interested in identifying words that associate with depression throughout the posting history, therefore, the bidirectional Long Short Term Memory (BiLSTM) of the original RCNN \cite{rcnn} implementation is replaced with Kim's \cite{https://doi.org/10.48550/arxiv.1408.5882} TextCNN. The TextCNN is implemented to learn the context representation from the stacked embedding vector, $\mathbf{c}_\textnormal{SE}$. The formulation of the CNN includes the convolution operation of a filter, which is applied to a window of $N_5$ words to produce a new feature. 

We have defined three filters 
$\mathbf{w} = \{1, 2, 3\}\in\mathbb R$,  
which are used to extract $N_5 \in$ \{unigram, bigram, and trigram\} of feature maps. 
A max-pooling layer is then applied to capture the most significant features from the generated feature maps, resulting in context 
$\mathbf{c}_\textnormal{CNN}$.

\begin{equation}
\mathbf{c}_{\textnormal{CNN}} = \textnormal{max} \{\mathbf{c}_{\mathbf{w_1}}, \mathbf{c}_{\mathbf{w_2}}, \mathbf{c}_{\mathbf{w_3}}\}
\end{equation}
Here, the context of the stacked embedding is reintroduced in the form of mean pooled,
\begin{equation}
\mathbf{c}_{\textnormal{SE}\_\textnormal{mean}} = \textnormal{mean} \{\mathbf{c}_{\textnormal{SE}}\}
\label{eq:se}
\end{equation}
and is then concatenated with the $\mathbf{c}_\textnormal{CNN}$, forming a rich SERCNN context $\mathbf{c}_\textnormal{SERCNN}$ as shown in Formula \ref{eq:sercnn}.

\begin{equation}
\mathbf{c}_{\textnormal{SERCNN}} = \{\mathbf{c}_{\textnormal{CNN}} \oplus \mathbf{c}_{\textnormal{SE}\_\textnormal{mean}}\}
\label{eq:sercnn}
\end{equation}
$\mathbf{c}_{\textnormal{SERCNN}}$ is then fed to two fully connected layers that have the following formulation,

\begin{equation}
o = W\mathbf{c} + b
\end{equation}
where $W$ is the trainable weight and $b$ is the bias term. The final output $\hat{y}$ presents as the classification output in probabilities using softmax:

\begin{equation}
     \hat{y}=softmax(o_2)
\end{equation}
where $o_2$ is the output of the second fully connected layer.

\subsection{Early Depression Detection}

From the statistics of our dataset (as tabulated in Table \ref{tab:tab-dataset-stat}), we found that not all users are active on social media. There are 75\% of users who have at least 49 posts, and the median of the dataset is 154 posts. This indicates that more than 50\% of the users have at least 154 posts (refer to Table \ref{tab:tab-dataset-stat}). Here, we propose to look into the viability of utilizing fixed observation windows, which are measured in the number of posts, to detect depression on social media. In this study, three observation windows: \{10, 30, 100\} posts are evaluated. 

\begin{figure}[htb!]
    \centering
    \includegraphics[width=0.4\linewidth]{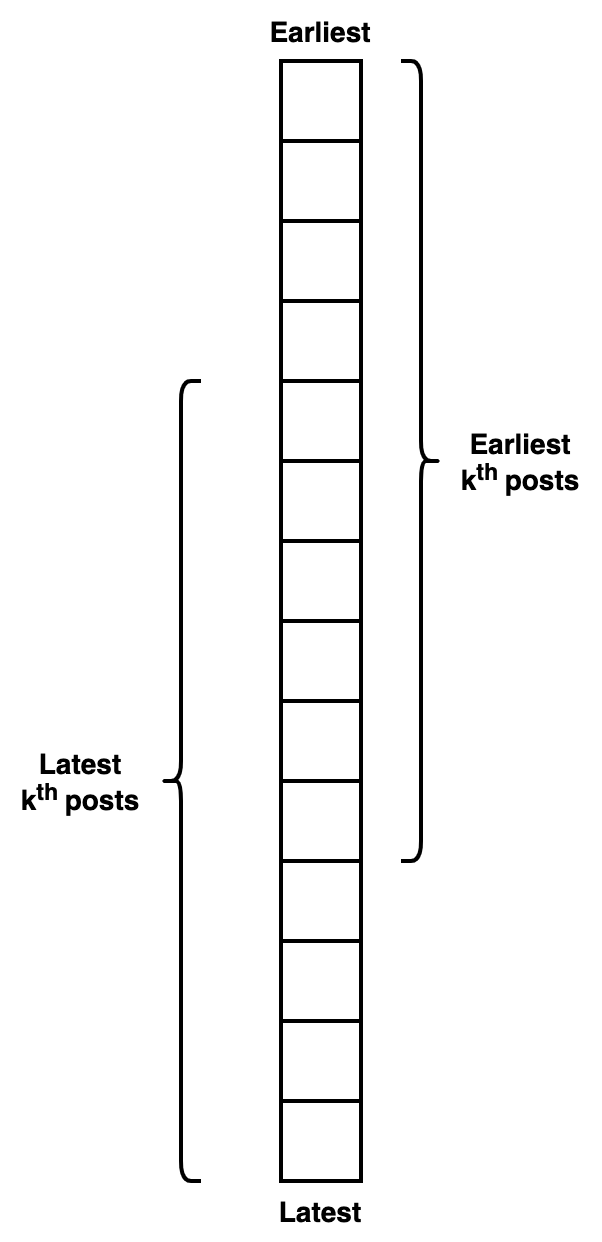}
    \caption{Assuming the user has 14 posts, model with the suffix -E is trained with the earliest $k^{th}$ posts while the one with the suffix -L is trained with the latest $k^{th}$ posts as illustrated.}
    \label{fig:obs-win}
\end{figure}

Since Shen et al.'s \cite{shen2017depression} dataset consists of a single month of posts before the ground truth tweet, we further assess the robustness of our proposed observation windows by training our model on two subsets of the posting history, denoted by a suffix of \textbf{-E} and \textbf{-L}.
Given that the number $k$ is one of the observation windows we experimented with, the model with the suffix -E is trained with the $k$ number of posts retrieved from the earliest $k$ posts. Similarly, models with the suffix -L are trained with the $k$ number of posts from the latest $k$ posts. An example is presented on Fig. \ref{fig:obs-win}. Do note that for some users with $k$ number of posts or less, the earliest $k^{th}$ and latest $k^{th}$ posts are the same, but this happens for several users only, as 75\% of the users have at least 49 posts and a median of 154 posts (refer to Table \ref{tab:tab-dataset-stat}). 

\section{Experiments}\label{sec:experiments}

\subsection{Preprocessing}\label{dataset} 

To ensure a fair comparison with previous papers, we do not expand the dataset and use the data included in the dataset only. The following preprocessing steps have been applied:
\begin{enumerate}
    \item removes ground truth (anchor tweet),
    \item lowercasing,
    \item removes URL,
    \item ensures only English healthy controls are included.
\end{enumerate}


\subsection{Experiment settings} 


We trained our model using the PyTorch library. All models are implemented with a cross-entropy function as the loss function and the overall objective function is optimized with the Adam \cite{adam} optimizer. Hyperparameters including learning rate, number of vocabulary, dropout, max training epochs, early stopping criteria, and batch size are reported in Table \ref{tab:parameter}. SERCNN and baseline models are evaluated using 5-fold cross-validation, with each fold of the train and test ratio being 4:1 and the train and validation ratio of 9:1.

\begin{table}[htbp!]
    \centering
    \caption{Hyperparameter settings} \label{tab:parameter}
    \begin{tabular}{lcc}
    \hline\hline
         {\textbf{Hyperparameter}}  & {\textbf{Value}}\\\hline\hline
         Learning rate & $1\mathrm{e}{-3}$\\
         Number of vocabulary & 10,000 \\
         Dropout  &  0.5 \\
         Max training epochs & 30\\
         Early stopping criteria & 10 \\
         Batch size & 120 \\
         \hline\hline
         
    \end{tabular}
\end{table}

\section{Results and Analysis}\label{sec:results}

We benchmark our SERCNN with our baseline models, as well as previous work that was trained on the same dataset \cite{gui2019depression,gui2019cooperative,shen2017depression}. Specifically, we compare SERCNN with LSTM, Hierarchical Attention Network \cite{yang-etal-2016-hierarchical}, Multimodal Dictionary Learning (MDL) and Multiple Social Networking Learning (MSNL) by \cite{shen2017depression} (which is trained with handcrafted features), CNN and LSTM with Policy Gradient Agent (PGA) by \cite{gui2019depression}, and Gated Recurrent Unit + VGG-Net + Cooperative Misoperation Multi-Agent (GVCOMMA) policy gradients by \cite{gui2019cooperative}. The performance of the models is evaluated with precision, macroaverage precision, macroaverage recall, and macroaverage F1 measure, which are tabulated in Table \ref{tab:rcnnvsbaseline}.

\begin{table}[htb!]
\centering
\caption{Performance comparison against baselines}
\label{tab:rcnnvsbaseline}
\begin{tabular}{lccccc}
\hline\hline
{\textbf{Model}}     & {\textbf{Features}}    & {\textbf{Accuracy}} & {\textbf{Precision}} & {\textbf{Recall}} & {\textbf{F1}} \\ \hline\hline
Shen et al.'s MSNL  \cite{shen2017depression}   & Handcrafted  & 0.818        & 0.818        & 0.818        & 0.818       \\
Shen et al.'s MDL \cite{shen2017depression}     &      features             & 0.848        & 0.848        & 0.85         & 0.849       \\ \hline
 
Gui et al.'s CNN + PGA  \cite{gui2019depression}  &    Text             & 0.871        & 0.871        & 0.871        & 0.871       \\
Gui et al.'s LSTM + PGA \cite{gui2019depression}    &                & 0.870         & 0.872        & 0.870        & 0.871       \\ \hline
Gui et al.'s GVCOMMA \cite{gui2019cooperative}    &      Text + image          & 0.900         & 0.900        & 0.901         & 0.900 \\ \hline


Our LSTM & Text & 0.900 & 0.921 & 0.900 & 0.910\\
Our HAN &  & 0.900 & 0.920 & 0.906 & 0.913\\
\textbf{Our SERCNN} &  & \textbf{0.937} & \textbf{0.929} & \textbf{0.941} & \textbf{0.933}\\\hline\hline
\end{tabular}
\end{table}

From Table \ref{tab:rcnnvsbaseline}, we can observe that the SERCNN outperformed all baseline models and previous works \cite{gui2019depression,gui2019cooperative,shen2017depression}. Overall, we have the following observations:

\begin{enumerate}
    \item The baseline LSTM model has better performance than Gui et al.'s LSTM variant \cite{gui2019depression}, which suggests that using pretrained embedding for depression detection on social media is an effective method for mitigating the effects of over-fitting using embedding trained from the dataset. 
    
    \item The performance of the baseline LSTM and HAN models is about the same indicating the hierarchical representation of the data does not necessarily represent the intrinsic structure of the data. 
    
    \item SERCNN outperforms other models, which can be due to the richer and more robust representation generated from the stacked embedding and the reintroduction of the embedding context, allowing the model to have a second chance to understand the context better.
    
\end{enumerate}


For early depression detection using our proposed observation windows, we have also trained a classifier with the pretrained \textit{bert-base-uncased} model (BertForSequenceClassification) from Hugging Face\footnote{\url{https://huggingface.co/}} \cite{wolf-etal-2020-transformers} as a baseline, since the concatenated text is much shorter now. The performance is presented in Table \ref{tab:compareresults2}.

\begin{table}[htb!]
\centering
\caption{Performance comparison using the earliest posts and the latest posts based on fixed observation windows.}
\label{tab:compareresults2}
\begin{tabular}{lccccc}\hline\hline
\textbf{Model} & \textbf{Number of posts} & \textbf{Accuracy} & \textbf{Precision} & \textbf{Recall} & \textbf{F1}\\\hline\hline

\textbf{BERT-E} & \textbf{10}&\textbf{0.878}&\textbf{0.872}&\textbf{0.882}&\textbf{0.873}\\
BERT-L & 10 &0.863&0.851&0.835&0.832\\
SERCNN-E & 10&	0.865&	0.858 &	0.871&	0.860 \\
SERCNN-L & {10} &0.870&	0.861 &	0.872 & 0.864 \\\hline

\textbf{BERT-E} & \textbf{30} &\textbf{0.907}&\textbf{0.902}& \textbf{0.909}&\textbf{0.902}\\
BERT-L & 30&0.903& 0.883&0.874& 0.872\\

 
\textbf{SERCNN-E} & \textbf{30} & \textbf{0.907}&\textbf{0.899} &	\textbf{0.910} & \textbf{0.903} \\
SERCNN-L & 30 &	0.900&	0.892&	0.902&	0.895 \\\hline

BERT-E & 100 &0.904&0.899& 0.904&0.899\\
BERT-L & 100 &0.901& 0.882&0.874& 0.871\\ 
\textbf{SERCNN-E} & \textbf{100} &	\textbf{0.922}&	\textbf{0.914}&\textbf{0.926}&	\textbf{0.918} \\
SERCNN-L & 100 &	0.915& 0.907&	0.919&	0.911 \\
\hline\hline
 
\end{tabular}
\end{table}

Based on the results shown in Table \ref{tab:compareresults2}, we observed that the BERT-E models for all observation windows performed better than their BERT-L, while the discrepancies between the SERCNNs are less than 1\% for all measured metrics. Comparing the performance of SERCNN with BERT, our proposed model shows competitive performance if not better than the finetuned BERT model despite using about a fraction of the number of parameters (about 2\%) as shown in Table \ref{tab:compare-parameters}. 

\begin{table}[htb!]
\centering
\caption{Comparison of BERT and SERCNN in the number of parameters}
\label{tab:compare-parameters}
\begin{tabular}{cc}
\hline\hline
\textbf{Model} & \textbf{Number of parameters} \\\hline\hline
BERT &  109,482,240 \\
SERCNN &  2,095,452\\ \hline\hline
\end{tabular}
\end{table}

In addition, these models also outperformed previous work that used the entire posting history. These findings suggest that depression symptoms are reflected in social media data throughout the month, which is consistent with the depression criteria stated in DSM-5 \cite{dsm5} where symptoms persist over time. Therefore, our empirical results indicate that not all posts in the user history are required to predict depression online.

\section{Conclusion} \label{sec:conclusion}

Empirical results have suggested that SERCNN has the advantage of achieving state-of-the-art accuracy and, at the same time, requires less computational cost and, most importantly, fewer posts for depression detection. This is crucial as different users may have different posting behaviors, ranging from less active to more active users. Concatenating social media posts into a single diary (document) allows the deep learning model to take advantage of the relationship of words in a different time frame, resulting in a generalized global context for a user. Adopting embeddings from general and social media domains allows better vocabulary coverage, thus reducing the out-of-vocabulary words for much more robust classification. Taking advantage of the rich representation, SERCNN achieved 93.7\% accuracy, outperforming previous works and baselines. Besides that, our experiment demonstrates that SERCNN can achieve 86.6\% accuracy using 10 posts only and 92.2\% accuracy using only 100 posts, instead of utilizing the whole dataset. We also show that SERCNN is neck-and-neck with fine-tuned BERT classifier across the three different observation window settings while having only 2\% of the BERT size in terms of parameter numbers. For our future work, we plan to explore incorporating transformer models into the architecture and looking into the transferability of the models on different datasets. 

\section*{Acknowledgment}

This work was carried out within the framework of the research project\\FRGS/1/2020/ICT02/MUSM/03/5 under the Fundamental Research Grant \\Scheme provided by the Ministry of Higher Education, Malaysia.

\section*{Ethics Statement}


Early screening of depression can significantly increase the discoverability of depression symptoms, allowing timely intervention to reduce symptoms. Our work provides novel insight into the potential of using a lesser number of posts that simulate the existing screening observation window of 14 days and reduce the cost of acquiring and storing data despite the limited number of real patient data and privacy concerns. However, our subjects are often sensitive and vulnerable; hence, additional measures have been taken to preserve their privacy. This study has received ethics approval from the \textit{Institutional Review Board} with review reference number: 2020-22906-40238.
Essential measures have been taken to exclude identifiable information from the data before it is used in this study. In addition, datasets obtained from their respective authors are stored in an encrypted repository. None of the users in the datasets is contacted throughout the study.
%
%
%
\bibliographystyle{splncs04}
\bibliography{references.bib}

\end{document}